# Bayesian Hierarchical Multi-Objective Optimization for Vehicle Parking Route Discovery

Romit S Beed

*Department of Computer Science, St. Xavier's College (Autonomous), Kolkata*

rbeed@yahoo.com

Sunita Sarkar

*Department of Computer Sc. & Engg., Assam University, Silchar.*

sunitasarkar@rediffmail.com

Arindam Roy

*Department of Computer Sc., Assam University, Silchar*

arindam_roy74@rediffmail.com

Abstract - Discovering an optimal route to the most feasible parking lot has been a matter of concern for any driver which aggravates further during peak hours of the day and at congested places leading to considerable wastage of time and fuel. This paper proposes a Bayesian hierarchical technique for obtaining the most optimal route to a parking lot. The route selection is based on conflicting objectives and hence the problem belongs to the domain of multi-objective optimization. A probabilistic data driven method has been used to overcome the inherent problem of weight selection in the popular weighted sum technique. The weights of these conflicting objectives have been refined using a Bayesian hierarchical model based on Multinomial and Dirichlet prior. Genetic algorithm has been used to obtain optimal solutions. Simulated data has been used to obtain routes which are in close agreement with real life situations.

*Keywords — Genetic Algorithm, Multi-objective, Optimization, Fitness, Weighted Sum Technique.*

## Introduction

Vehicle Routing Problems (VRPs) are popular combinatorial optimization problems present in transport sector which generally include allotting resources and scheduling in constrained environments from a depot to different locations having considerable financial implications. VRPs have gained popularity in the recent past because of the its diverse applicability and commercial importance in defining effective distribution tactics to shrink operative costs in distribution networks. A typical VRP involves designing minimum cost routes from a principal warehouse to a set of physically distributed locations having varying requirements. Each such location needs to be serviced precisely once by a single van, and each van has a fixed carrying capacity. Vehicle parking route determination may be considered as an extension of a typical VRP. Instead of considering only distance as a parameter for determining the most appropriate parking lot, the time it takes to reach the particular lot and the availability of empty space may also be conflicting factors while determining the most suitable lot. Hence this problem may be treated as a multi-objective optimization problem.

A multi-objective optimization problem (MOP) deals with two or more objectives or parameters contributing towards the final output, and in general these objectives influence one another in a complicated and nonlinear manner [1]. The main goal is to obtain a set of values for these objectives, thereby generating an optimization of the overall problem. There has been an extensive use of evolutionary computation in solving MOPs [2–5] as they provide high quality solution through proper mapping of genetic information and fitness calculation. According to Goldberg, multi-objective or multi-criteria optimization is the process of optimizing multiple conflicting objectives in parallel subject to a set of constraints. In such problems, it is seen that there is not an individual solution that concurrently minimizes each objective to the fullest, but to a limit exceeding which the additional objective(s) will be compromised as a result [6]. After obtaining a particular solution, one of the main aims of MOPs is to compare it with other solutions and assess how improved this solution is in comparison to the existing set of solutions [7]. A multi-objective problem including several, conflicting objectives may be formulated into a one-objective scalar function. This popular technique, known as weighted-sum method or Single Objective Evolutionary Algorithm (SOEA), is an a priori technique based on "linear aggregation of functions" principle [8]. By definition, the weighted-sum method reduces to a positively weighted convex sum of the objectives, as follows:

$$\min \sum_{i=1}^{n} w_i \cdot f_i(x), where \sum_{i=1}^{n} w_i = 1; w_i > 0 \forall i \qquad (1)$$

Minimization of this single-objective function is expected to give an efficient solution for the original multi-objective problem. The process involves scalarizing the conflicting objectives into a single objective function. There are various scalarization techniques which have been proposed in the past. Zadeh popularized the weighted sum technique as a classical approach for solving such problems [9]. This method, as the name suggests, scalarizes a set of conflicting objective functions, by pre-multiplying each of the objective function by predefined weights. Though the technique is straightforward and computationally efficient, it may not be able to investigate all solutions when the true Pareto front in non-convex [10]. Multinomial distribution, being the multivariate generalization of binomial distribution, is used to model the probability of more than two quantitatively specified mutually exclusive outcomes occurring in a set of repeated trials and hence is a discrete multivariate distribution. Dirichlet distribution is multivariate generalization of beta and it is well known that Dirichlet distribution acts as a conjugate prior for Multinomial distribution.

Majority of the existing works have aimed at choosing the set of weights such that it stabilizes the solution set [11-13], this work proposes to frame a model which determines a much stable set of weights in comparison to that obtained deterministically. The criticisms of the existing methodologies for determination of weights have motivated this work and to propose the Bayesian model based on multinomial and Dirichlet priors. As per the authors' existing knowledge, this work is first of its kind since none of the earlier works had this motivation of searching for stability in weights. The model has been so developed as to reflect the relative importance of the conflicting objectives through the respective weights, which were stochastically estimated, based on the data obtained from a pilot survey for the given purpose. Unlike the available continuation techniques, this method can be applied with convenience to handle any number of objectives. As this method yields a posterior probability distribution over the weights, the stochastically generated weight vectors can be used to obtain solutions with less computational complications.

## Literature Review

The Vehicle Routing Problem dates back to 1950s. Dantzig and Ramser aimed to design a solution for delivering gasoline to petrol pump using a mathematical programming model. Ever since then, VRP has gained popularity and evolved substantially and is a subject of research in varied domains. Obtaining the most optimal route thereby reducing the total transportation expense is the main objective of this approach. However, in reality the modeling is much more complicated than the traditional VRP [14]. Thus this traditional VRP model has evolved over time, constraints have been added like vehicle capacity and the model has been named the Capacitated Vehicle Routing Problem (CVRP); another constraint is the time interval in which each customer has to be served named as the Vehicle Routing Problem with Time Windows (VRPTW). Other variations of VRP which exist are multiple depot VRP, periodic VRP, split delivery VRP, stochastic VRP, VRP with backhauls, VRP with pickup and delivering to name a few. Vehicle Routing Problem with Time Windows (VRPTW) is an improvised VRP where a time window is defined for each user. Apart from the vehicle size limitation, each user defines a time frame within which an individual service has to be accomplished, e.g. packing or unpacking of items, parking a vehicle etc. It may be possible that a particular vehicle may arrive before the scheduled time, in that case it will have to wait till the scheduled start of service time.

Yan Han et al suggested the use of WiFi positioning technology [15] to resolve the parking problem by designing a technique of programmed assignment based on the personal demands of the users and avoidance of traffic conflicts. The major decision factors which were taken into consideration were lane usage conditions, driving distance, walking distance, and the occupancy rate of parking space. Optimal routes were obtained using the Djikstra's algorithm; this route was then conveyed to the driver using the mobile network. Compared to traditional GPS positioning and mobile cellular network locating, WiFi positioning technology provided better results thus leading to more effective use of the limited parking resources. The Siemens Integrated Smart Parking Solution [16] is a traffic coordination system that streamlines driver decision-making using heuristic data about parking space obtainability. Drivers are directed to the assigned parking location through the most optimum route using the in-car navigation system. RFID technique is used along with sensors to monitor, track and store data of every car involved in the system. The key features of the system are (i) Dynamic guidance which helps reduce manual search thereby lessens the traffic; (ii) Improve route selection from source to destination using statistical data based on multiple objectives; (iii) Gather Knowledge and make available parking price and regulations as well as enforce parking violations fines cost-effectively by accessing real-time data; (iv) Improve resource utilization by dynamically rerouting vehicles to optimize the usage of the parking lots at different points of the city

Selection of accurate weights can lead to better performance of an algorithm. Timothy Ward Athan [17] proposed a quasi-random weighted criteria system that produces weights covering the Pareto set consistently. The method is based on random probability distribution and involves a large number of computations. Gennert and Yuille [18] proposed a nonlinear weight determination algorithm where an optimal point is obtained that is not in the vicinity of the extreme points. Although a lot of work is available in the literature regarding systematic selection of the weights in solving a Multi-Objective Optimization problem, till date a comprehensive data driven technique determining weights reflecting the relative importance of the conflicting objectives is lacking. Most of the work in the available literature has focused on fixing the weights based on some prior beliefs or information In some of the existing literature wi's have been estimated simply by the proportion of preference in the respective categories. Pertaining to the above mentioned concerns regarding selection od weights, this work has been based on the weights generated according to a hierarchical Bayesian technique which generates the weights based on data derived from a pilot survey.

Genetic Algorithm (GA) is a metaheuristic based on the concept of natural selection, designed to obtain, produce or choose a heuristic that provides a sufficiently favorable solution to an optimization problem, particularly with inadequate or partial information or restricted computing capacity. GAs fall under the category of Evolutionary Algorithms (EAs) and are used to obtain superior results to optimization and search requirements by basing on bio-inspired mechanisms like selection, crossover and mutation. It aims at generating the most optimal result to a problem within a vast set of likely possible results through genetic enhancement across generations. John Holland [19] and his team at the University of Michigan presented an algorithm that modified a population of individual entities, each having a corresponding fitness value, into an evolved generation based on the Darwinian principle of evolution and policy of 'survival of the fittest'. Bryant et al. [20] suggested that Genetic Algorithms may be used in optimizing systems based on the natural phenomenon of evolution. The concept of survival of the fittest may be incorporated in designing an algorithm which discovers or explores, though not all possible results, and obtains a fair result. The process starts with an initial guess value, and gradually generates the fittest solutions through several generations of iterations whereby the result of each generation is better than the result of the earlier generation.

GAs generates a population of solutions using the popular crossover and mutation operators. Chand et al [21] proposed new methods for genetic operators - Sub Route Mapped Crossover Method (SMCM) and Sub Route Exchange Mutation Method (SEMM). They used Dominant Rank method to obtain Pareto Optimal Set and the two objectives that they considered were number of vehicles and total cost (distance). The Dominant Rank Method found optimum solutions efficiently. As GAs provide ease of operation and possess widespread applicability, it is one of the most trusted algorithms in computational optimization and operations research. A noble optimization technique requires balancing the degree of exploration of information gathered till the present generation through the processes of recombination and mutation with the degree of exploitation through the selection technique [22]. If the results gathered are exploited exceedingly, early convergence is evident. Conversely, if the search is extensive, it shows that the information gathered till then has not been used properly. By changing the parameters of the GA operators, exploration and exploitation can be controlled. Thus Genetic Algorithms have become the most extensively used tool to solve the vehicle routing problem.

# Multi Objective Optimization Model

Multi-objective optimization is becoming widespread in the current technology driven optimization scenarios. Typical examples of multiple-objectives are reduction in cost, increase of performance, growth of reliability, reduction of time, shortening of travel distance, maximization of speed etc. It can be observed that many of these objectives may be conflicting with one another, and optimizing the problem with respect to one objective may generate undesired results with respect to the other objectives. A better approach to such a multi-objective problem is to discover a number of solutions, each of which satisfies the objectives up to a particular permissible level without being dominated by any other result. The two most common approaches for solving multiple-objective optimization are the utility theory method or the weighted sum method and the Pareto optimal method. The weighted sum approach combines the different objective functions into a single function and then obtains an optimal solution. The drawback that remains with this approach is in deciding the weights of the different objectives. The Pareto optimal approach on the other hand, obtains a set of values that are non-dominated with respect to each other. The weighted sum technique is generally employed to transform multiple objectives into one objective. This is implemented by issuing different weights to the various objectives which are chosen based on the objective's relative significance with respect to the other objectives. The single objective optimization aims at minimizing the overall cost as follows:

$$\min z = w_1 z_1(x) + w_2 z_2(x) + \ldots + w_k z_k(x), \qquad (2)$$

where $z_i(x)$ are the normalized objective functions and $\sum w_i = 1$.

Developing a model to yield an optimal route to the lot involves multiple conflicting objectives. Here we have explored the weighted sum technique to scalarize these objectives into a single objective function. However, the weighted sum method suffers from the problem of scientific determination of weights for the conflicting objectives. This work is based on the weights generated by a novel Bayesian hierarchical method using Dirichlet distribution. The choice of weights is guided by a probability model which helps us draw data based inference. To decide upon the weights, reflecting the relative importance of the objectives, a pilot survey is conducted. The number of individuals opting for the various mutually exclusive categories of conflicting objectives is noted as $n_1$, $n_2$ and $n_3$. Assuming (n1, n2, n3) ~ Multinomial distribution with parameters (n; $w_1, w_2, w_3$), where $w_i$ (weight corresponding to the i$^{th}$ objective), is the proportion of individuals in the population (unknown) who will vote for category i of the objective function or is the probability that an individual selected randomly from the population votes for i$^s$ category. Assuming ($w_1$, $w_2$, $w_3$) to be distributed as Dirichlet with concentration parameters $a_1, a_2$ and $a_3$, The estimates of weights are taken as

$$w_i = \frac{a_i + n_i}{\sum_{i=1}^{3}(a_i + n_i)}, \; i = 1,2,3 \qquad (3)$$

where $a_1, a_2$ and $a_3$ are the values maximizing the marginal likelihood function of data given $a_1, a_2$ and $a_3$.

This paper presents an innovative fitness function that models the real life situation in the most appropriate way. A part of a city map has been modelled. Two locations '0' and '1' (marked in red) are considered to be the start locations. There are 26 intermediate junctions, '2'-'27' (marked in blue) and there are three parking lots, '28'-'30' (marked in green). The optimality and suitability of the parking lot as well as the path leading to it from the source is obtained using a fitness function that is formulated based on three factors - distance, speed and parking availability. Google Map API has been used to obtain the distance between these points as well as the average speed of journey in the same route.

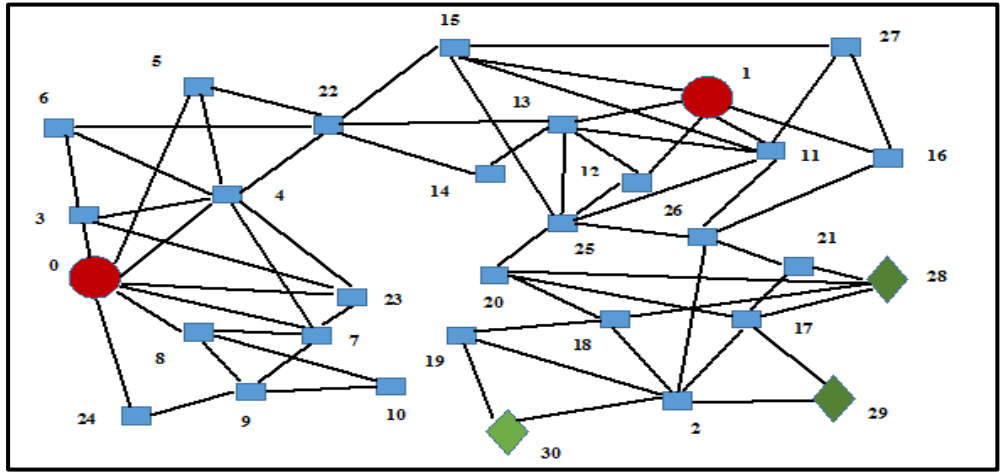

Fig 1: Road map

The formulation of the minimization function of three conflicting objectives is as follows:

$$\min f = \min(f_1, f_2, f_3) \quad (4)$$

The **first objective function** minimization of distance function $f_1$ is formulated as follows:

$$f_1 = \sum_{i,j \in N} d(i,j) \quad (5)$$

Where N= set of all nodes

The **second objective function** is maximization of speed $f_2$ is defined as follows:

$$f_2 = -\sum_{i,j \in N} s(i,j) \quad (6)$$

where $s_{ij}$ is the speed form node i to node j, which may vary over different segments of the path. It is known that maximization(f) ≡ minimization (-f).

The **third objective function** is maximization of the parking availability and is defined as follows:

$$f_3 = -x_p \quad (7)$$

where $x_p$ represents the availability of the parking lot 'p'.

**Constraints**

$$d_{i,j} \geq 0 \; \forall \; i,j \in N; \; s_{i,j} \geq 0 \; \forall \; i,j \in N; \; x_p \geq 0 \; \forall \; p \in N$$

The fitness function is modeled as the weighted sum of the above defined objective functions as follows:

$$\text{Min } f = w_1 f_1 + w_2 f_2 + w_3 f_3 \quad (8)$$

where $\sum_{i=1}^{3} w_i = 1$

A pilot survey was conducted among 50 drivers. The result of the survey was as follows:

| Description | Count | Frequentist Weight | Bayesian Weight | Err Var Frequentist | Err Var Bayesian |
|---|---|---|---|---|---|
| Total participants in the survey: | 50 | 1.00 | 1.00 | | |
| Highest priority to distance to parking lot: | 16 | 0.32 | 0.29 | 0.00230 | .00044 |
| Highest priority to travel speed on the route: | 14 | 0.28 | 0.30 | 0.00300 | .00056 |
| Highest priority to parking availability at parking lot: | 20 | 0.40 | 0.41 | .00330 | .00062 |

Table 1: Weights - Frequentist and Bayesian

The weights generated by the Bayesian Hierarchical method in (3) have been found to be much stable with respect to the error variance as compared to the classical approach. In fact the improvement in error variance under small sample sizes is almost 100% by this method.

The paper aims to provide a solution using Genetic Algorithm to solve Multi-Objective optimization problem. There are three factors which are considered, the distance, the speed and availability of the parking lot. The key components of GAs are: initial population of chromosomes, selection based on the fitness value, crossover to produce better descendants and mutation to produce randomness. The fitness function allocates a fitness value to every chromosome in the existing population. The fitter and better chromosomes are selected based on higher fitness values which are then selected for the Crossover. Crossover involves combining the bits of one chromosome with the other chromosome to produce a child chromosome that inherits characteristics of both the father and the mother. Mutation is then performed to retain genetic diversity across generations. One such iteration is termed as a generation. Based on the principle of survival of the fittest, the algorithm controls which chromosomes should survive, die or reproduce.

The proposed algorithm is as follows:
*Step 1: Generate randomly an initial population of N chromosomes.*
*Step 2: Compute the fitness value of all chromosomes using the fitness function.*
*Step 3: Perform tournament selection by selecting individuals from the population and insert into the mating pool.*
*Step 4: Perform single point crossover with a crossover rate of 0.2*
*Step 5: Perform Creep Mutation where a random gene is selected and its value is changed with a random value between lower and upper bound.*
*Step 6: Update population.*
*Step 7: If termination condition is met, then stop else go to Step 2.*

**Initial Population**
The first step involves producing an initial population. Each chromosome consists of a sequence of nodes starting from the start node, through one or multiple intermediate nodes terminating in the parking lot node. Every gene of the chromosome comprises of a node identifier ensuring that the node does not reappear. A typical chromosome is represented as follows:
**[0, 5, 22, 15, 25, 26, 2, 19, 30]**
where {0} is the starting node and marked red in figure 1, {30} is the parking lot marked green in the figure 1 and {5}, {22}, {15}, {25}, {26}, {2} and {19} are the intermediate nodes marked blue in figure 1.

**Selection (Tournament Selection)**
An efficient and robust selection mechanism generally used by genetic algorithms is Tournament Selection. The procedure involves selecting fitter chromosomes and inserting into a mating pool. Chromosomes from this mating pool are then made to reproduce and generate new offspring for the next generation. It is necessary that the mating pool contains "high quality" chromosomes. The technique conducts a tournament among S competitors, where S is called the tournament size. The chromosome with the best fitness value within the S competitors is selected as the winner and shifted to the mating pool which stores all the tournament winners. Considering a tournament size S=3, three chromosomes are randomly selected from the population. fitter among them is chosen as one of the parents. For example
Chromosome 1:  [0, 6, 22, 15, 25, 20, 18, 28]                      fitness= .728542
Chromosome 2:  [0, 8, 7, 23, 3, 6, 22, 13, 25, 11, 2, 29]          fitness= .859741
Chromosome 3:  [0, 4, 22, 14, 13, 25, 11, 26, 2, 19, 17, 20, 18, 28]  fitness = .958308
Chromosome 1, being the fittest among the three gets selected as Parent 1. Since the problem deals with minimization, lower the fitness value, fitter is the chromosome.
Chromosome 4:  [0, 4, 22, 14, 13, 25, 11, 26, 2, 19, 17, 20, 18, 28]  fitness= .958308
Chromosome 5:  [0, 7, 4, 6, 22, 13, 1, 11, 2, 17, 21, 28]          fitness= .960787
Chromosome 6:  [0, 3, 6, 22, 15, 1, 11, 26, 21, 28]                fitness= . 827978
Here, Chromosome 6 is fittest and is selected as Parent 2. Thus Chromosome 1 and Chromosome 6 are selected as Parent 1 and Parent 2 for crossover. The complexity of tournament selection is O(n).

**Crossover (Single Point Crossover)**
The next step involves performing the crossover operation and the performance of GA significantly depends on this crossover operators. New generation of chromosomes are created from the existing generation by crossing over the genetic information of the parents to produce a fitter child. Various crossover techniques exist that obtain optimum results over the generations. Single Point Crossover has been used here where the two mating chromosomes are bisected at a single point and the halves are interchanged. An integer m is assumed for each parent chromosome such that m < n, where n is the length of the chromosome. The chromosomes are then bisected at the m$^{th}$ point and genetic information is interchanged. Considering m to be the midpoint of the length, each parent contributes 50% of its characteristics to the child.

Suppose there are the following two parents:
Parent 1 (Chromosome 1): [0, 6, 22, 15, 25, 20, 18, 28]
Parent 2 (Chromosome 6): [0, 3, 6, 22, 15, 1, 11, 26, 21, 28]
The length of parent P$_1$ is 8 so it contributes the first 50% of itself (first 4 genes) to the crossover process. The length of parent P$_2$ is 10 so it contributes the second half of itself (last 5 genes) to the crossover process. In this way the two children are created.
Child 1: [0, 6, 22, 15, 1, 11, 26, 21, 28]
Child 2: [0, 3, 6, 22, 15, 25, 20, 18, 28]
The fitter of the two children is then shortlisted for the mutation process.

**Mutation (Creep Mutation)**
Mutation is a genetic operator implemented to provide genetic variety from one generation to the next by simply modifying one or multiple genes. Similar to crossover, there are various mutation techniques. Creep Mutation has been used her where an arbitrary gene is chosen and its information is replaced with another random information within the bounds. The rate of mutation has been set as 1/n where n is the length of the chromosome i.e. only one gene in the entire chromosome is mutated. The gene to be mutated can be only replaced by a gene which will produce a valid path. Considering that child 2 had been selected since it had a better fitness value than child 1, the input to the mutation process is as follows:
Child 2 (before mutation): [0, 3, 6, 22, 15, 25, 20, 18, 28]
A gene is selected at random, say {3}. This gene can be replaced by {4} since a valid path can be generated from {0} to {6} through {4}. Hence, the new chromosome is as follows:
Child 2 (after mutation): [0, 4, 6, 22, 15, 25, 20, 18, 28]
Transforming the population with crossover and mutation operator will usually take O(NL) where N is the size of the population L is the length of the chromosome.

**Normalization of Objectives**
One has to normalize the objectives while formulating the composite objective function using the weighted sum method. As distance is in the range of kilometers, speed is in the range of kilometers/hours and availability is a percentage value, there is this need to normalize them in the range [0,1] to bring them down to a common range for optimization. It is best to normalize them to a range [0,1]. In order to scale a range $[a, b]$ to [0,1] the following function is used:

$$f(x) = \frac{x-a}{b-a} \qquad (9)$$

# Results and Discussion

Initially, the classical weighted sum technique has been implemented on 30 locations, numbered from 1 to 30 for reference. The distance matrix and the speed matrix are [30] x [30] matrices generated at run time by using the Google Maps API. The following three APIs have been used (a) Google Maps Distance Matrix API, (b) Google Maps Geocoding API and (c) Google Maps Places API. Thirty generations have been considered. The fitness values for the different epochs have been tabulated. It is observed that as the generations increases, the value of the fitness function tend to decrease till it stabilizes at an optimal value.

| Gen | 12-4 am | 4-8 am | 8-12 pm | 12-4 pm | 4-8pm | 8-12am | Gen | 12-4 am | 4-8 am | 8-12 pm | 12-4 pm | 4-8pm | 8-12am |
|---|---|---|---|---|---|---|---|---|---|---|---|---|---|
| 1 | 0.729 | 0.664 | 0.598 | 0.789 | 0.633 | 0.564 | 16 | 0.577 | 0.568 | 0.465 | 0.678 | 0.538 | 0.476 |
| 2 | 0.719 | 0.664 | 0.598 | 0.789 | 0.633 | 0.564 | 17 | 0.566 | 0.568 | 0.465 | 0.654 | 0.538 | 0.458 |
| 3 | 0.696 | 0.656 | 0.592 | 0.750 | 0.633 | 0.531 | 18 | 0.561 | 0.553 | 0.450 | 0.654 | 0.538 | 0.458 |
| 4 | 0.696 | 0.656 | 0.592 | 0.750 | 0.619 | 0.531 | 19 | 0.554 | 0.549 | 0.440 | 0.643 | 0.538 | 0.458 |
| 5 | 0.675 | 0.632 | 0.586 | 0.750 | 0.598 | 0.517 | 20 | 0.550 | 0.522 | 0.440 | 0.632 | 0.512 | 0.427 |
| 6 | 0.672 | 0.632 | 0.568 | 0.732 | 0.566 | 0.499 | 21 | 0.548 | 0.509 | 0.438 | 0.632 | 0.512 | 0.427 |
| 7 | 0.672 | 0.629 | 0.554 | 0.726 | 0.566 | 0.499 | 22 | 0.533 | 0.498 | 0.438 | 0.632 | 0.512 | 0.413 |

| 8 | 0.630 | 0.625 | 0.539 | 0.709 | 0.566 | 0.499 | 23 | 0.533 | 0.498 | 0.434 | 0.615 | 0.512 | 0.413 |
|---|---|---|---|---|---|---|---|---|---|---|---|---|---|
| 9 | 0.630 | 0.608 | 0.523 | 0.709 | 0.566 | 0.476 | 24 | 0.518 | 0.476 | 0.434 | 0.608 | 0.512 | 0.413 |
| 10 | 0.610 | 0.608 | 0.523 | 0.697 | 0.566 | 0.476 | 25 | 0.498 | 0.476 | 0.428 | 0.608 | 0.512 | 0.413 |
| 11 | 0.599 | 0.587 | 0.509 | 0.691 | 0.541 | 0.476 | 26 | 0.498 | 0.472 | 0.425 | 0.596 | 0.512 | 0.404 |
| 12 | 0.599 | 0.587 | 0.495 | 0.678 | 0.538 | 0.476 | 27 | 0.485 | 0.472 | 0.425 | 0.596 | 0.512 | 0.404 |
| 13 | 0.597 | 0.587 | 0.489 | 0.678 | 0.538 | 0.476 | 28 | 0.485 | 0.472 | 0.425 | 0.596 | 0.512 | 0.404 |
| 14 | 0.597 | 0.587 | 0.479 | 0.678 | 0.538 | 0.476 | 29 | 0.485 | 0.472 | 0.425 | 0.596 | 0.512 | 0.404 |
| 15 | 0.596 | 0.572 | 0.465 | 0.678 | 0.538 | 0.476 | 30 | 0.485 | 0.472 | 0.425 | 0.596 | 0.512 | 0.404 |

Table 2: Hour zone wise fitness data across Generations

A plot has been made of fitness vs generation of the above values.

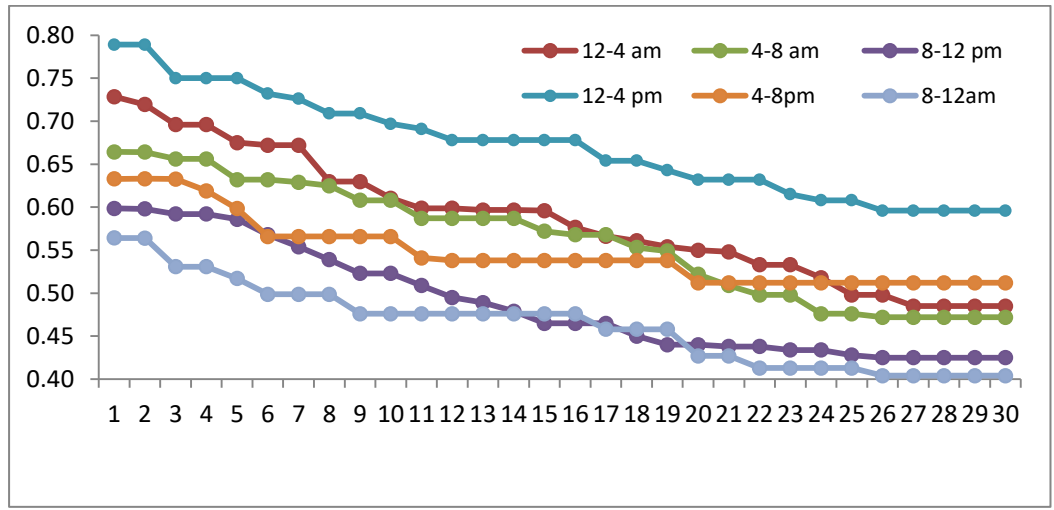

Figure 2: Plot of Fitness vs. Generations for different time slots

| TIME SLOT | ROUTES |
|---|---|
| 12am- 4am | [0, 4, 22, 14, 13, 11, 26, 21, 28] |
| 4am-8am | [0, 4, 22, 13, 11, 2, 29] |
| 8am-12pm | [0, 3, 4, 22, 13, 25, 20, 28] |
| 12pm-4pm | [0, 5, 22, 13, 11, 2, 29] |
| 4pm-8pm | [0, 7, 4, 22, 13, 25, 20, 28] |
| 8pm-12am | [0, 5, 22, 13, 11, 2, 17, 21, 28] |

Table 3: Route table

Thus from the above graph and its associated route table (Table 2) it is seen that the fitness value always exhibits a consistently decreasing trend as the number of generations increase across all time zones. Secondly the routes as well as the parking lot vary depending on the time zone. This simulates a real life scenario where parking lots and routes are bound to change as the values for the different factors changes. Although distance remains constant but the average speed and availability of parking lots changes with time which gets finally reflected in the fitness function. On further analyzing the data to understand the effect of each of the parameters - distance, speed and parking availability, the following was observed.

**12am-4am:** During these hours the roads remain empty, the parking lots remain available. The average speed on all major roads is almost the same. So in this epoch distance has a higher priority over speed and parking lot availability. Routes with shorter distances were selected in this time zone.

**4am-8am:** This marks a transition period. The later hours of this epoch mark the beginning of the peak hours or office going hours of the day. So parking lots start filling up slowly. Distance still has the highest priority here followed by speed and parking availability. There was no marked domination by any one parameter.

**8am-12pm:** This marks the peak hours of the day. The roads remain congested. Parking lots start filling up fast. Thus now speed has the highest priority followed by parking availability, followed by distance. Fastest routes featured in the result set more often showing the dominance of the speed parameter.

**12pm-4pm:** During this time the parking lots are filled up. Congestion on roads reduces to some extent. As a result parking lot availability has the highest priority here followed by speed and then distance. Parking availability was the key parameter for route selection in this time zone.

**4pm-8pm:** This marks the time when people start returning to their houses. Parking lots start becoming empty. Here speed and distance more or less have the same priority. Parking lot availability has the least priority here. There was no marked observation of a single parameter dominating others.

**8pm-12am:** This marks the night time. Roads start becoming empty, parking lots are almost empty, so again distance gains the maximum priority followed by speed and then parking availability. Distance again became the dominating parameter in this time zone.

# Conclusion

The work has intended to implement a route discovery technique based on Bayesian hierarchical concept aimed at promoting sustainable development in a developing nation. Development of smart cities aids nations to improve their environmental conditions and help in maintaining a greener and cleaner world. This work relied on sound statistical techniques to improve the route optimization process for sustainable development through conservation of resources. The focus was on the improvement of the weights representing the relative importance of the possibly conflicting objective functions under consideration by proposing a flexible Hierarchical Bayesian model. This technique has been analysed for error variances thereby quantifying the reliability of the estimates. Bayesian determination of weights finds high applicability in cases where conducting a large scale survey is time consuming, difficult to implement as well as expensive. When applied in the domain of route optimization in discovering the most suitable parking lot, the proposed methodology have produced results which display close resemblance to the phenomenon observed in real life situations. If implemented in reality, this would certainly ensure saving of time, energy and fuel, thus a greener world.

The Weighted Sum Method has been used here to solve the problem using Genetic Algorithm. This work can be expanded further by considering Pareto optimal solutions by plotting the Pareto curve and obtaining solutions from the Pareto front. The system may be improved by varying the various GA parameters such as population size, number of generations and mutation rate. A prototype model designed above has helped us to study the impact of each of the conflicting objectives' dominance at different hours of the day and how their priorities change over time. The system has been implemented with thirty one locations which can be increased to a bigger domain. The system may be further improved to fetch and work with dynamic data reflecting the changing objective parameters during different hours of the day.

References

[1] Beatrice Ombuki, Brian J. Ross And Franklin Hanshar, Multi-Objective Genetic Algorithms for Vehicle Routing Problem with Time Windows Applied Intelligence 24, Springer Science + Business Media, Inc. 17–30, 2006
[2] C.A. Coello Coello, D.A. Van Veldhuizen, and G.B. Lamont, EvolutionaryAlgorithms for Solving Multi-Objective Problems, Kluwer Academic Publishers, 2002
[3] C. M. Fonseca and P. J. Fleming. "An overview of evolutionary algorithms in multiobjective optimization," Evolutionary Computation, vol. 3, no. 1, pp. 1–16, 1995.
[4] D. A. Van Veldhuizen and G. B. Lamont, "Multiobjective evolutionary algorithms: Analyzing the state-of-the-art," Evolutionary Computation, vol. 8, no. 2, pp. 125–147, 2000.
[5] K. Deb, Multi-objective Optimization using Evolutionary Algorithms, John Wiley, 2001.
[6] Goldberg, D.E. 1998 Genetic Algorithm in Search, Optimization, and Machine Learning. Addison-Wesley, Reading, Massachusetts.
[7] Haupt, R.L., Haupt, H.S. 2004 Practical Genetic Algorithms. 2nd Edition, A John Wiley & Sons, Inc., Publication.
[8] R Arulmozhiyal and A M Jubril, "A nonlinear weights selection in weighted sum for convex multi-objective optimization". FACTA UNIVERSITATIS (NI_S) Ser. Math. Inform. Vol. 27 No 3, 2012, pp 357-372.
[9] L A Zadeh, "Optimality & non-scalar-valued performance criteria". IEEE Trans. Automat. Control, AC-8, 1963, 59–60.
[10] Abdullah Konak et al., Multi-objective optimization using genetic algorithms: A tutorial, Reliability Engineering and System Safety 91 (2006) 992–1007.
[11] T W Athan & P Y Papalambros, "A note on weighted criteria methods for compromise solutions in multiobjective optimization", Eng. Opt., 27, 1996 pp. 155–176.
[12] J H Ryu, S. Kim, H. Wan, "Pareto front approximation with adaptive weighted sum method in multi-objective simulation optimization" In Proceedings of the 2009 Winter Simulation Conference, 2009, pp 623–633.

[13] D Schmaranzer, R Braune, & K F Doerner, Multi-objective simulation optimization for complex urban mass rapid transit systems, Annals of Operations Research, 2019, pp 1–38.

[14] Vehicle Routing Problem, Edited by Tonci Caric and Hrvoje Gold, Published by In-The, Croatian branch of I-Tech Education and Publishing KG, Vienna, Austria., ISBN 978-953-7619-09-1.

[15] Yan Han, Jiawen Shan, Meng Wang, Guang Yang, Optimization design and evaluation of parking route based on automatic assignment mechanism of parking lot , Advances in Mechanical Engineering 2017, Vol. 9(7) 1–9, https://doi.org/10.1177/1687814017712416

[16] Smart City Vision – Movement through Innovation, www.siemens.com/ mobility/smart-parking

[17] T W Athan, P Y Papalambros, "A quasi-montecarlo method for multi-criteria optimization". Eng. Opt., 27 1996, pp 177–198.

[18] M A Gennert, A L Yuille, "Determining the optimal weights in multiple objective function optimization". In Second International conference on computer vision, IEEE, Los Alamos, CA, September 1998, pp 87–89.

[19] J. H. Holland, Adaptation in natural and artificial systems. Ann Arbor, MI: University of Michigan Press, 1975.

[20] K. Bryant, "Genetic algorithms and the traveling salesman problem." Master's thesis, Harvey Mudd College, Claremont, United States, 2000.

[21] Padmabati Chand, J. R. Mohanty, A Multi-objective Vehicle ,Routing Problem using Dominant Rank Method, International Conference in Distributed Computing & Internet Technology (ICDCIT-2013) Proceedings published in International Journal of Computer Applications (IJCA) (0975 – 8887) pp 29 -34

[22] Martin Pelikan, Genetic Algorithms, MEDAL Report No. 2010007, 2010.